\documentclass{article}
\usepackage{ijcai22}

\usepackage[numbers,sort&compress]{natbib}

\usepackage{times}
\usepackage{soul}
\usepackage{url}
\usepackage[hidelinks]{hyperref}
\usepackage[utf8]{inputenc}
\usepackage[small]{caption}
\usepackage{graphicx}
\usepackage{amsmath}
\usepackage{amsthm}
\usepackage{booktabs}
\usepackage{algorithm}
\usepackage{algorithmic}
\usepackage{cleveref}

\usepackage[usenames,dvipsnames]{xcolor}

\definecolor{personal}{HTML}{F24726}
\definecolor{contextual}{HTML}{9510AC}
\definecolor{dynamic}{HTML}{0CA789}
\definecolor{instruction}{HTML}{414BB2}
\definecolor{evaluation}{HTML}{F24726}

\newcounter{rqcount}
\newcommand{\researchquestion}[1]{\begin{quote}RQ\therqcount: \emph{#1}\end{quote}\stepcounter{rqcount}}

\everypar{\clubpenalty=0}
\everypar{\widowpenalty=0}
\everypar{\displaywidowpenalty=0}
\everypar{\looseness=-1}
\title{Humans are not Boltzmann Distributions: Challenges and Opportunities\\ for Modelling Human Feedback and Interaction in Reinforcement Learning}

\author{
David Lindner$^1$
\And
Mennatallah El-Assady$^1$
\affiliations
$^1$ ETH Zurich
\emails
david.lindner@inf.ethz.ch,
menna.elassady@ai.ethz.ch
}

\begin{document}

\maketitle

\begin{abstract}
Reinforcement learning (RL) commonly assumes access to  well-specified reward functions, which many practical applications do not provide. Instead, recently, more work has explored learning what to do from interacting with humans. So far, most of these approaches model humans as being (nosily) rational and, in particular, giving unbiased feedback. We argue that these models are too simplistic and that RL researchers need to develop more realistic human models to design and evaluate their algorithms. In particular, we argue that human models have to be \emph{personal}, \emph{contextual}, and \emph{dynamic}. This paper calls for research from different disciplines to address key questions about how humans provide feedback to AIs and how we can build more robust human-in-the-loop RL systems.
\end{abstract}

\section{Introduction}

Reinforcement learning (RL) has been  successful in solving complex tasks, such as playing video games \cite{mnih2015human} or controlling robotic systems \cite{haarnoja2018learning}. However, it remains challenging to apply RL to tasks without a well-specified reward function, such as autonomous driving \cite{knox2021reward}. RL from human feedback is a promising alternative that aims to interactively learn from human feedback instead of a fixed reward function \cite{christiano2017deep}.

Unfortunately, existing approaches to learning from human feedback rely on simple human models, such as  Boltzmann distributions \cite{jeon2020reward}. We argue that currently used models are too simplistic. We discuss challenges we expect such methods to encounter as they are being increasingly deployed in practical applications and present opportunities for further research towards improved modeling of human feedback.

In this paper, we, first, a brief overview of commonly used types of human-AI interaction and work in RL using human feedback (\Cref{sec:related_work}). Then, we present a range of open research questions about modeling human feedback (\Cref{sec:modelling_human_feedback}) and key dimensions for designing RL systems that can robustly learn from humans (\Cref{sec:rl_implications}).  We raise many research questions about human modeling that are highly relevant to any human-AI interaction. We close by discussing the implications of our analysis and a call for interdisciplinary research addressing these questions (\Cref{sec:discussion}).

\paragraph{Position Statement.} We argue that human-in-the-loop reinforcement learning needs more realistic human models. 

\paragraph{Contributions.} We, propose a framework for modeling human feedback as \emph{personal}, \emph{contextual}, and \emph{dynamic}. For each category, we discuss open research questions both from a social science  and a computer science perspective, and call for interdisciplinary research to address these questions.

\section{Related Work}\label{sec:related_work}

Research on human-centered machine learning has borrowed many methods from social science and the humanities to assess human feedback and interaction. However, most of the assumptions made about the reasoning processes, interaction and communication dynamics, as well as the integration of such feedback into learning models has remained unchallenged, lacking effective human-centered evaluations~\cite{sperrle2021survey}.

In this section, we provide a brief overview of types of human-AI interaction, with a specific focus on RL.

\subsection{Collaborative Human-AI Interaction}\label{sec:human_feedback}

\begin{figure}
    \centering
    \includegraphics[width=0.9\linewidth]{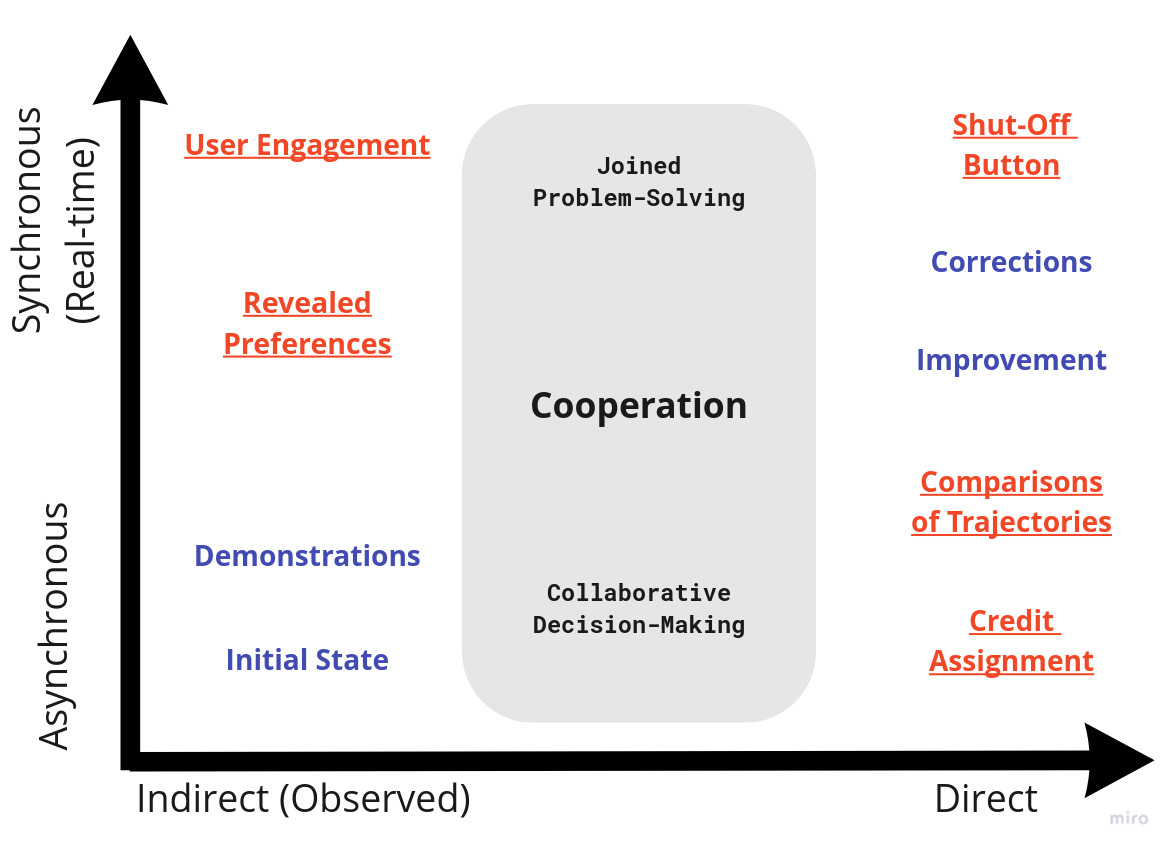}
    \vspace{-1em}
    \caption{
        We classify a subset of common types of human feedback in RL along three dimensions. Humans can give \emph{direct} feedback to an RL agent performing a task. The agent can also obtain \emph{indirect} feedback by observing the human. Feedback can be provided \emph{synchronously}, i.e., in real-time, or \emph{asynchronously}, i.e., before or after the agent acts. Finally, feedback is {\color{instruction}\textbf{instructive}} if it tells the agent what to do explicitly; feedback is {\color{evaluation}\textbf{\underline{evaluative}}} if it only tells the agent how good it is doing.
    }
    \label{fig:types_of_feedback}
\end{figure}

To give an overview of the most common types of human-AI interaction, we provide a categorization in \Cref{fig:types_of_feedback}. We distinguish three categories of interaction: \emph{Instruction}, \emph{Evaluation}, and \emph{Cooperation}.
For a more thorough review of possible types of human-AI interaction in the context of RL, we refer to recent reviews \cite{lin2020review,jeon2020reward}.

\paragraph{Instruction.}
Human feedback is \emph{instructive} if it tells the agent \emph{what to do}. For example, demonstrations show the agent how to do a specific task. In some situations, the agent can also indirectly obtain information about what to do simply by observing the state of the world~\cite{shah2019preferences}. More directly, the human can tell the agent what to do by \emph{correcting} it physically~\cite{bajcsy2017learning}, or providing \emph{improvements}, e.g., an alternative action in a specific situation~\cite{jain2015learning}.

\paragraph{Evaluation.} Human feedback is \emph{evaluative} if it tells the agent \emph{how well it is doing}. Humans can provide this information directly, e.g., by comparing trajectories~\cite{christiano2017deep} or shutting off the agent \cite{hadfield2017off}. The agent can also get implicit evaluations, e.g., by measuring user engagement~\cite{zhao2018explicit}, or by monitoring gestures or facial expressions~\cite{cui2020empathic}.

\paragraph{Cooperation.} More complicated forms of human-AI interaction need to be modeled in a cooperation framework~\cite{hadfield2016cooperative}. While we could also model all instructive and evaluative feedback as a form of cooperation, the latter allows for more general forms of interaction beyond giving feedback. For example, a human and an RL agent might have to solve a problem together, requiring them to learn from each other~\cite{carroll2019utility}.

\subsection{Models of Human Feedback in RL}

The most popular way to integrate human feedback in RL is learning from demonstrations, where the agent learns a task from observing a (human) expert's \emph{demonstrations} of a task. This can happen either via imitation learning \cite{hussein2017imitation}, or via inverse reinforcement learning \cite{ng2000algorithms}. The main limitation of this approach is that the demonstrations have to be (close to) optimal, which is often difficult for humans to achieve. Instead of demonstrations, a human can also provide a direct \emph{reinforcement} signal, e.g., a binary rating ``good'' or ``bad'' \cite{knox2009interactively}. In many applications, it can be easier to provide contrastive feedback, e.g., by comparing two possible actions or trajectories, rather than an absolute evaluation. This motivates \emph{preference-based} RL \cite{wirth2017survey}.

All of these methods require a model of how humans give feedback. They commonly assume that humans are perfectly rational or at least unbiased. One of the most common ways to model noisy feedback is using a Boltzmann distribution \cite{jeon2020reward}. Some work tries to evaluate the effect of this \textit{misspecification} \cite{freedman2021choice}. Other work tries to learn biases in human feedback from data, with mixed results \cite{shah2019feasibility}. Overall, current methods in RL from human feedback are heavily affected if their assumptions about the human model are not satisfied \cite{lee2021b}.

The B-Pref benchmark \cite{lee2021b} models \textit{irrationalities} that humans might exhibit when giving feedback, intending to move towards more realistic human models. The benchmark considers humans making systematic mistakes, skipping comparisons of very similar or very different trajectories, or giving myopic feedback, i.e., weighting recently seen things higher when making decisions. 
We consider this work a promising step in the right direction. However, none of the simulations in B-Pref is grounded in the existing literature on cognitive biases, and the benchmark is not based on human data.

Some literature on recommender systems has explored the use of more advanced human models, but these methods are often tailored to the specific applications and have not made it to more general applications of RL yet \cite{pu2011user,jameson2015human}.

\section{Challenges in Human Feedback Modelling} 
\label{sec:modelling_human_feedback}

Most existing approaches to incorporating human feedback into RL, assume humans act noisily but are unbiased. It is common to assume that humans are goal-driven and act rationally consistent. This model is, of course, wrong, and humans are far from rational and unbiased.

We argue that in practice, human-AI interaction is \emph{personal}, \emph{contextual}, and \emph{dynamic}. %
This section discusses these three aspects, focusing on open research questions. The next section will discuss potential implications for designing systems that interactively learn from humans.

Most of our discussion applies to any human-AI interaction. However, to be concrete, we focus on RL agents receiving evaluative or instructive feedback. This setting inherits most problems of modeling humans in general, but it allows for more concrete research on practical applications.

\subsection{\textit{Personalized} Feedback \&  Interaction}\label{sec:personalized_feedback}

We cannot hope to find a universal model that describes how humans interact with RL systems that we can use to design these systems. Human-AI interaction is inherently personal, and we need to model it as such.

How a human interacts with an RL agent will depend on personality factors~\cite{nunes2012personality}. For example, it is likely that in many tasks, the user's conscientiousness will affect how they evaluate an agent. A crucial first step towards designing personalized interaction is to understand how personality affects human-AI interactions.

In addition to different personalities, each person also has prior knowledge they bring into an interaction. Whether the user is a domain expert, an ML expert, or a novice, will affect which situations they can judge and which types of feedback they can give reliably. For example, an expert might be able to provide near-optimal demonstrations, whereas a novice might not. Still, the novice might be able to evaluate the task competently when giving the right user interface~\cite{dudley2018review}.

As a first step towards building personalized human models, we propose to study the following research questions:

\researchquestion{How do \textbf{personality factors} influence how humans interact with AI systems?}

\researchquestion{Are there \textbf{measurable} personality factors with a clear impact on the human feedback and interaction dynamic to allow for personalization?}

\researchquestion{How can we \textbf{quantify} prior knowledge and semantic understanding to model interaction dynamics?}

\subsection{\textit{Contextualized} Feedback \&  Interaction}\label{sec:contextualized_feedback}

To model human-AI interaction accurately, we need to take into account the (sociotechnical) \emph{context} of the interaction. In particular, the interaction dynamics depend on when and where it is happening and which social and cultural norms exist around the interaction~\cite{ehsan2021explainable,ehsan2020human}.
For example, a medical doctor might be more careful about evaluating the information from an AI system than the average user of a personal smartphone assistant.
In the context of RL the context of the interaction depends on the environment in which the agent and the human act, and how the human relates to this environment.

To enable contextualized modeling of human-AI interaction, we need to understand:

\researchquestion{Which aspects of the \textbf{sociotechnical context} affect human-AI interaction?}

\researchquestion{How do \textbf{cultural} (and other) \textbf{differences} influence the interaction?}

\researchquestion{How do contextual factors influence individual personality factors, and can they be modeled \textbf{independently}?}

\subsection{\textit{Adaptive} Feedback \&  Interaction}\label{sec:adaptive_feedback}

A major limitation of most human models used in the RL literature is that they are static, i.e., they do not change throughout the interaction. This is unrealistic; in practice, both the user and the AI system will accumulate knowledge that changes how they interact with each other~\cite{sperrle2021co}. Other factors, such as the user's energy and motivation levels, might change over time and affect the interaction. Moreover, the context of the interaction might change due to external factors.

RL systems have to adapt to these \emph{dynamic} factors and adapt how they interact with humans over time. To build robost adaptive systems, it is necessary to investigate:

\researchquestion{Which factors have temporal and interaction-dependent \textbf{variation}?}

\researchquestion{Can we \textbf{measure} and \textbf{predict} changes in these factors during interaction sequences?}

\researchquestion{How do personal and contextual factors \textbf{adapt} to changes in the interaction?}

\begin{figure}
    \centering
    \includegraphics[width=0.8\linewidth]{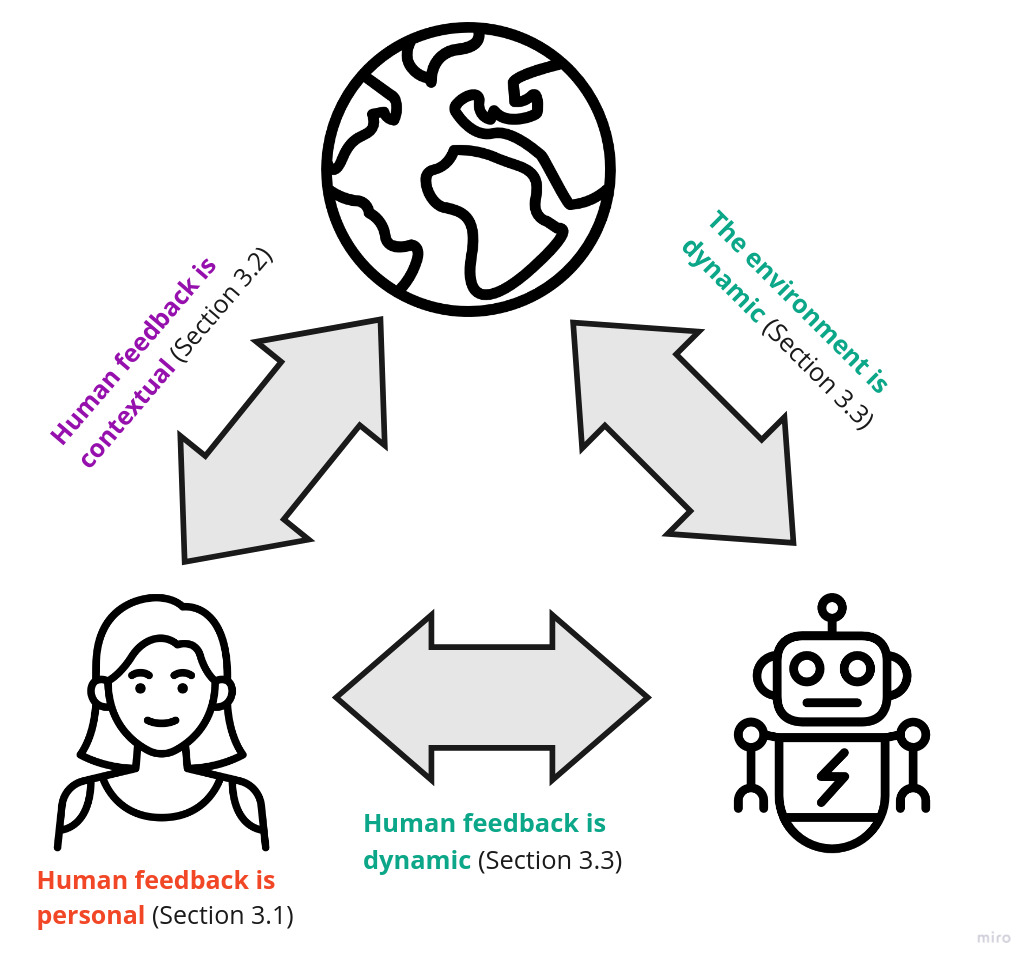} \vspace{-1em}
    \caption{Schematic overview of RL with human feedback. In \Cref{sec:modelling_human_feedback}, we discuss the key components of the interaction. \Cref{sec:rl_implications} discusses key dimensions for robust interactive learning: {\color{personal}\textbf{Personalized Learning}} (\Cref{sec:personalized_learning}), {\color{contextual}\textbf{Contextualized Modeling}} (\Cref{sec:contextualized_learning}), and {\color{dynamic}\textbf{Adaptive Querying}} (\Cref{sec:adaptive_querying}).}
    \label{fig:overview}
\end{figure}

\section{Implications for RL Applications}
\label{sec:rl_implications}

Now, let us turn our focus towards designing RL systems that learn from interacting with humans. Along our three dimensions of modelling human feedback and interaction, we now highlight how to design RL systems that are personalized, contextualized and adaptive.

\subsection{\textit{Personalized} Learning}
\label{sec:personalized_learning}

To build RL systems that can learn from different people, we need to ensure that all parts of the interaction are personalized. Building on a better understanding of the aspects discussed in \Cref{sec:personalized_feedback}, we could decide which system designs are most appropriate for which users. This includes answering the following research questions:

\researchquestion{How should we choose the \textbf{type} of human feedback for an individual user?}

\researchquestion{How much can an RL system \textbf{trust} the responses of an individual user?}

\researchquestion{Which \textbf{explanations} can an RL system provide to an individual user to allow them to give better feedback?}

\subsection{\textit{Contextualized} Modeling}
\label{sec:contextualized_learning}

Similar to building personalized learning systems, we want to ensure to make them aware of the sociotechnical context of the interaction as well. Building on the questions discussed in \Cref{sec:contextualized_feedback}, we want to answer similar questions from an RL perspective:

\researchquestion{In which context can an RL system \textbf{rely} on users providing high-quality feedback?}

\researchquestion{How can an RL system provide \textbf{context-appropriate} explanations of it's behavior?}

\researchquestion{How can we ensure RL systems can learn from human feedback reliably even in \textbf{novel} contexts?}

\subsection{\textit{Adaptive} Querying}
\label{sec:adaptive_querying}

We need to enable RL systems to adapt to changes in their interaction with humans. This is particularly important for designing human-in-the-loop systems that do not have a separate phase of learning from humans, but continuously interact with humans during their deployment. Such systems need to have uncertainty estimates that are well-calibrated with respect to how users and interaction patterns might change. We propose several concrete research questions in this direction:

\researchquestion{How can RL systems make \textbf{situation-aware} queries to users?}

\researchquestion{How can RL systems maintain appropriate \textbf{uncertainty} to \textbf{detect} changes in the interaction?}

\researchquestion{How much can systems learn about a user's interaction patterns \textbf{online}, and how many inductive biases do we need to \textbf{encode}?}

\section{Discussion}\label{sec:discussion}

After discussing the most important dimensions for designing more human-centered RL systems, let us highlight the key challenges and research opportunities.

\subsection{Challenges}

\paragraph{An Interdisciplinary Research Challenge.} Modeling human feedback is not \emph{only} a technical problem. The most crucial open research questions need to be answered from a human-centered perspective. However, no single research discipline is currently offering insights on the right level of granularity. On the one hand, research in Neuroscience aims to provide a detailed understanding on the basis of human behavior. However, its insights are not (yet) actionable for our purpose. On the other hand, behavioral and cognitive psychology research asks many similar questions. However, its focus is often too broad to yield actionable insights on how to model human feedback in RL systems.

\paragraph{More Difficulty for RL.} The main reason for using simple human models in current work is that they make learning more tractable. If we can no longer get unbiased feedback about rewards, how can we hope to still learn a good RL policy. Developing more realistic human models will likely make RL more difficult.

\subsection{Research Opportunities}

\paragraph{An Interdisciplinary Approach.} To make progress in modeling human feedback, we need to answer multi- and interdisciplinary research questions. These will require research from a wide range of disciplines to engage with these questions, including, but not limited to, researchers from Computer Science, (Cognitive) Psychology, Ethics, Philosophy, Behavioral Science, Communication Sciences, Sociology, and Neuroscience. To answer the research questions we posed, we need to first compare the current understanding from different disciplines. Then, we can aim to operationalize the open questions within specific disciplines and design experiments to answer them.

\paragraph{Towards More Robust RL.} Ultimately, we hope to gain insights into measurable factors that affect human-AI interaction, which can be used to build better human models. Such models promise to make it easier to design new RL algorithms that learn from humans and benchmark existing methods in more realistic simulations without running user studies to evaluate small algorithmic changes.

\paragraph{An Opportunity for New Algorithms.} A human-centered perspective on the RL learning problem creates challenges for applying existing methods, that assume access to a reward function. But, it provides an opportunity for developing novel, human-centered learning algorithms that are designed with the goal of learning from and with humans.

\section{Conclusion}

We reviewed research on human feedback in RL from a human-centered perspective. We argued that current human models used in RL are too simple and that we need better human models if we want to design systems that can learn from humans robustly in the real world. We argued that we need personal, contextual, and dynamic models to design robust RL systems that learn from humans. We hope to start an interdisciplinary discussion around these topics with the goal of building better human models and designing interaction protocols that can work outside of simulations.

\bibliographystyle{plain}
\bibliography{references}

\end{document}